\documentclass[10pt, a4paper]{article}
\usepackage{lrec}
\usepackage{multibib}
\newcites{languageresource}{Language Resources}
\usepackage{graphicx}
\usepackage{tabularx}
\usepackage{soul}
\usepackage{amsmath}
\usepackage{upquote}
\usepackage{epstopdf}
\usepackage[latin1]{inputenc}

\usepackage{hyperref}
\usepackage{xstring}

\title{SentEval: An Evaluation Toolkit for Universal Sentence Representations}

\name{Alexis Conneau$^*$\thanks{$^*$LIUM, Universit\'e Le Mans} and Douwe Kiela}
\address{Facebook Artificial Intelligence Research \\
         \{aconneau,dkiela\}@fb.com\\}

\abstract{
We introduce SentEval, a toolkit for evaluating the quality of universal sentence representations. SentEval encompasses a variety of tasks, including binary and multi-class classification, natural language inference and sentence similarity. The set of tasks was selected based on what appears to be the community consensus regarding the appropriate evaluations for universal sentence representations. The toolkit comes with scripts to download and preprocess datasets, and an easy interface to evaluate sentence encoders. The aim is to provide a fairer, less cumbersome and more centralized way for evaluating sentence representations.\\\
\newline  }

\begin{document}

\maketitleabstract

\newcommand{\insertclassiftasks}{
    \begin{table*}[h!]
        {\small
        \centering
        \begin{tabular}{|l|r|l|c|p{8.6cm}|c|}
        \hline \bf name & \bf N & \bf task & \bf C & \bf examples & \bf label(s) \\
        \hline
        MR & 11k &  sentiment (movies) & 2 &  ``Too slow for a younger crowd , too shallow for an older one.'' & neg\\
        CR & 4k &  product reviews & 2 & ``We tried it out christmas night and it worked great .'' & pos\\
        SUBJ & 10k & subjectivity/objectivity & 2 &  ``A movie that doesn't aim too high , but doesn't need to.'' & subj \\
        MPQA & 11k &  opinion polarity & 2 & ``don't want'';  ``would like to tell''; & neg, pos\\
        TREC & 6k & question-type & 6 &  ``What are the twin cities ?'' & LOC:city\\
        SST-2 & 70k &  sentiment (movies) &  2 & ``Audrey Tautou has a knack for picking roles that magnify her [..]'' & pos \\
        SST-5 & 12k &  sentiment (movies) &  5 & ``nothing about this movie works.'' & 0 \\
        \hline
        \end{tabular}
        \caption{\label{table:classif} {\bf Classification tasks}. C is the number of classes and N is the number of samples.}
        }%
    \end{table*}
}

\newcommand{\insertpairtasks}{
    \begin{table*}[h!]
        {\small
        \centering
        \begin{tabular}{|l|r|l|c|p{4.6cm}|p{4.6cm}|c|}
        \hline \bf name & \bf N & \bf task & \bf output & \bf premise & \bf hypothesis & \bf label \\
        \hline
        SNLI & 560k & NLI & 3 & ``A small girl wearing a pink jacket is riding on a carousel.'' & ``The carousel is moving.'' & entailment\\ 
        SICK-E & 10k & NLI & 3 & ``A man is sitting on a chair and rubbing his eyes'' &  ``There is no man sitting on a chair and rubbing his eyes'' & contradiction\\
        SICK-R & 10k & STS & $[0, 5]$ & ``A man is singing a song and playing the guitar'' & ``A man is opening a package that contains headphones'' & 1.6\\
        STS14 & 4.5k & STS & $[0, 5]$ & ``Liquid ammonia leak kills 15 in Shanghai'' & ``Liquid ammonia leak kills at least 15 in Shanghai'' & 4.6\\
        MRPC & 5.7k & PD & 2 & ``The procedure is generally performed in the second or third trimester.'' & ``The technique is used during the second and, occasionally, third trimester of pregnancy.'' & paraphrase\\
        COCO & 565k & ICR & sim & \raisebox{-0.5\totalheight}{\includegraphics[width=0.26\textwidth, height=18mm]{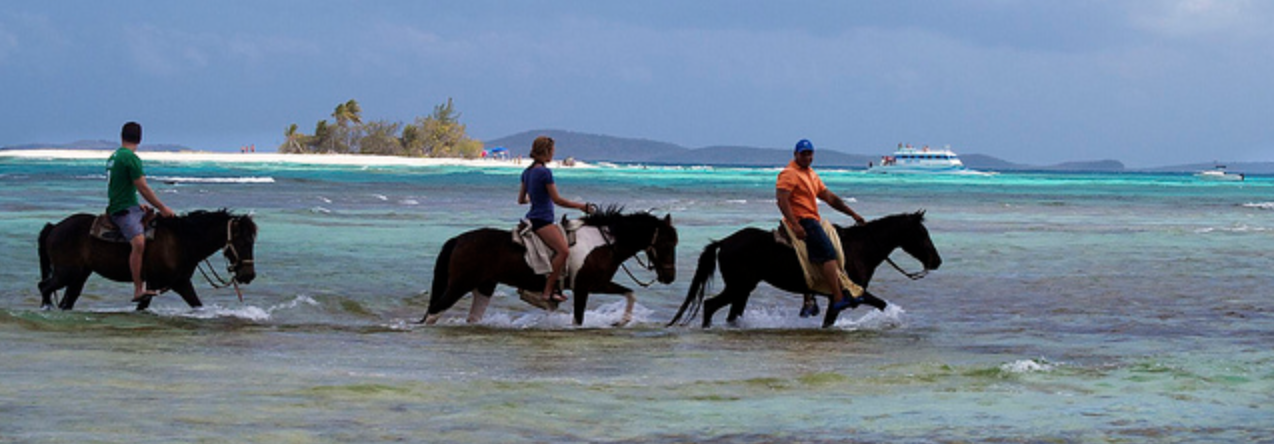}}\footnotemark{}  & ``A group of people on some horses riding through the beach.'' & rank\\
        & & & & & & \\
        \hline
        \end{tabular}
        \caption{\label{table:sts} {\bf Natural Language Inference and Semantic Similarity tasks}. NLI labels are contradiction, neutral and entailment. STS labels are scores between 0 and 5. PD=paraphrase detection, ICR=image-caption retrieval.}
        }%
    \end{table*}
}

\newcommand{\inserttransferclassif}{
    \begin{table*}[t]
        \resizebox{0.99\linewidth}{!}{
        \begin{tabular}{@{}l@{\,}|ccccccccc}
        \hline
        \bf Model & \bf MR & \bf CR & \bf SUBJ & \bf MPQA & \bf SST-2 & \bf SST-5 & \bf TREC & \bf MRPC &  \bf SICK-E  \\
        \hline
        \hline
        \multicolumn{9}{l}{\it Representation learning (transfer)} & \\
        \hline
        GloVe LogReg & 77.4 &78.7 & 91.2 & 87.7 & 80.3 & 44.7 & 83.0 & 72.7/81.0 & 78.5  \\
        GloVe MLP & 77.7 & 79.9 & 92.2 & 88.7 & 82.3 & 45.4 & 85.2 & 73.0/80.9 & 79.0 \\
        fastText LogReg & 78.2 & 80.2 & 91.8 & 88.0 & 82.3 & 45.1 & 83.4 & 74.4/82.4 & 78.9 \\
        fastText MLP & 78.0 & 81.4 & 92.9 & 88.5 & 84.0 & 45.1 & 85.6 & 74.4/82.3 & 80.2 \\
        SkipThought & 79.4 & 83.1 & 93.7 & 89.3 & 82.9 & - & 88.4 & 72.4/81.6 & 79.5  \\
        InferSent & 81.1 & 86.3 & 92.4 & 90.2 & 84.6 & 46.3 & 88.2 & 76.2/83.1 & 86.3  \\
        % 81.7   86.4   92.5   90.4   85.1   45.8   88.2   76.5   86.4
        \hline
        \hline
        \multicolumn{9}{l}{\it Supervised methods directly trained for each task (no transfer)} & \\
        \hline
        SOTA & 83.1$^1$ & 86.3$^1$ & 95.5$^1$ & 93.3$^1$ & 89.5$^2$ & 52.4$^2$ & 96.1$^2$ & 80.4/85.9$^3$ & 84.5$^4$  \\
        \hline
        \end{tabular}
        }
        \caption{Transfer test results for various baseline methods. We include supervised results trained directly on each task (no transfer). Results $^1$ correspond to AdaSent \protect\cite{zhao2015self}, $^2$ to BLSTM-2DCNN \protect\cite{zhou2016text}, $^3$ to TF-KLD \protect\cite{ji2013discriminative} and $^4$ to Illinois-LH system \protect\cite{lai2014illinois}.
        \label{table:classif_results}
        }
    \end{table*}
}

\newcommand{\inserttransferrelated}{
    \begin{table*}[h!]
        \centering
        \resizebox{0.85\linewidth}{!}{
        \begin{tabular}{@{}l@{\,}|ccccccc}
        \hline  \bf Model & \bf SST'12 & \bf SST'13 & \bf SST'14 & \bf SST'15 & \bf SST'16 & \bf SICK-R & \bf SST-B \\
        \hline
        \hline
        \multicolumn{8}{l}{\it Representation learning (transfer)} \\
        \hline
        GloVe BoW & 52.1 & 49.6 & 54.6 & 56.1 & 51.4 & 79.9 & 64.7  \\
        fastText BoW & 58.3 & 57.9 & 64.9 & 67.6 & 64.3 & 82.0 & 70.2  \\
        SkipThought-LN & 30.8 & 24.8 & 31.4 & 31.0 & - & 85.8 & 72.1  \\
        InferSent & 59.2 & 58.9 & 69.6 & 71.3 & 71.5 & 88.3 & 75.6  \\
        Char-phrase & 66.1 & 57.2 & 74.7 & 76.1 & - & - & - \\
        \hline
        \hline
        \multicolumn{8}{l}{\it Supervised methods directly trained for each task (no transfer)} \\
        \hline
        PP-Proj & 60.0$^1$ & 56.8$^1$ & 71.3$^1$ & 74.8$^1$ & - & 86.8$^2$ & - \\
        
        \end{tabular}
        }
        \caption{Evaluation of sentence representations on the semantic textual similarity benchmarks. Numbers reported are Pearson correlations x100. We use the average of Pearson correlations for STS'12 to STS'16 which are composed of several subtasks. Charagram-phrase numbers were taken from \protect\cite{wieting2016charagram}. Results $^1$ correspond to PP-Proj \protect\cite{wieting2015towards} and $^2$ from Tree-LSTM \protect\cite{tai2015improved}.
        \label{table:related_results}
        }
    \end{table*}
}

\section{Introduction}
\label{sec_introduction}

Following the recent word embedding upheaval, one of NLP's next challenges has become the hunt for universal general-purpose sentence representations. What distinguishes these representations, or embeddings, is that they are not necessarily trained to perform well on one specific task. Rather, their value lies in their transferability, i.e., their ability to capture information that can be of use in any kind of system or pipeline, on a variety of tasks.

\-\hspace{0.1cm} Word embeddings are particularly useful in cases where there is limited training data, leading to sparsity and poor vocabulary coverage, which in turn lead to poor generalization capabilities. Similarly, sentence embeddings (which are often built on top of word embeddings) can be used to further increase generalization capabilities, composing unseen combinations of words and encoding grammatical constructions that are not present in the task-specific training data. Hence, high-quality universal sentence representations are highly desirable for a variety of downstream NLP tasks.

\-\hspace{0.1cm} The evaluation of general-purpose word and sentence embeddings has been problematic \cite{Chiu:2016repeval,Faruqui:2016repeval}, leading to much discussion about the best way to go about it\footnote{See also recent workshops on evaluating representations for NLP, e.g. RepEval: https://repeval2017.github.io/}. On the one hand, people have measured performance on intrinsic evaluations, e.g. of human judgments of word or sentence similarity ratings  \cite{Agirre:2012semeval,Hill:2016cl} or of word associations \cite{Vulic:2017eacl}. On the other hand, it has been argued that the focus should be on downstream tasks where these representations would actually be applied \cite{Ettinger:2016acl,Nayak:2016acl}. In the case of sentence representations, there is a wide variety of evaluations available, many from before the ``embedding era'', that can be used to assess representational quality on that particular task. Over the years, something of a consensus has been established, mostly based on the evaluations in seminal papers such as SkipThought \cite{Kiros:2015nips}, concerning what evaluations to use. Recent works in which various alternative sentence encoders are compared use a similar set of tasks \cite{Hill:2016naacl,Conneau:2017emnlp}.

%\-\hspace{0.1cm} 
Implementing pipelines for this large set of evaluations, each with its own peculiarities, is cumbersome and induces unnecessary wheel reinventions. Another well-known problem with the current status quo, where everyone uses their own evaluation pipeline, is that different preprocessing schemes, evaluation architectures and hyperparameters are used. The datasets are often small, meaning that minor differences in evaluation setup may lead to very different outcomes, which implies that results reported in papers are not always fully comparable.

\-\hspace{0.1cm} In order to overcome these issues, we introduce SentEval\footnote{https://github.com/facebookresearch/SentEval}: a toolkit that makes it easy to evaluate universal sentence representation encoders on a large set of evaluation tasks that has been established by community consensus.
\section{Aims}
\label{sec_aims}

The aim of SentEval is to make research on universal sentence representations fairer, less cumbersome and more centralized. To achieve this goal, SentEval encompasses the following:

\begin{itemize}
\item one central set of evaluations, based on what appears to be community consensus;
\item one common evaluation pipeline with fixed standard hyperparameters, apart from those tuned on validation sets, in order to avoid discrepancies in reported results; and
\item easy access for anyone, meaning: a straightforward interface in Python, and scripts necessary to download and preprocess the relevant datasets.
\end{itemize}

In addition, we provide examples of models, such as a simple bag-of-words model. These could potentially also be used to extrinsically evaluate the quality of word embeddings in NLP tasks.

\insertclassiftasks

\insertpairtasks
\section{Evaluations}
\label{sec_evaluations}

Our aim is to obtain general-purpose sentence embeddings that capture generic information, which should be useful for a broad set of tasks. To evaluate the quality of these representations, we use them as features in various transfer tasks.

\paragraph{Binary and multi-class classification}
We use a set of binary classification tasks (see Table~\ref{table:classif}) that covers various types of sentence classification, including sentiment analysis 
(MR and both binary and fine-grained SST) \cite{Pang:2005acl,Socher:2013emnlp}, question-type (TREC) \cite{Voorhees:2000sigir}, product reviews (CR) \cite{Hu:2004kdd}, subjectivity/objectivity (SUBJ) \cite{Pang:2004acl} and opinion polarity (MPQA) \cite{Wiebe:2005lrec}. We generate sentence vectors and classifier on top, either in the form of a Logistic Regression or an MLP. For MR, CR, SUBJ and MPQA, we use nested 10-fold cross-validation, for TREC cross-validation and for SST standard validation.

\footnotetext{Antonio Rivera - CC BY 2.0 - flickr}

\paragraph{Entailment and semantic relatedness}
We also include the SICK dataset \cite{Marelli:2014lrec} for entailment (SICK-E), and semantic relatedness datasets including SICK-R and the STS Benchmark dataset \cite{cer2017semeval}. For semantic relatedness, which consists of predicting a semantic score between 0 and 5 from two input sentences, we follow the approach of \newcite{Tai:2015acl} and learn to predict the probability distribution of relatedness scores. SentEval reports Pearson and Spearman correlation. In addition, we include the SNLI dataset \cite{Bowman:2015emnlp}, a collection of 570k human-written English supporting the task of natural language inference (NLI), also known as recognizing textual entailment (RTE) which consists of predicting whether two input sentences are entailed, neutral or contradictory. SNLI was specifically designed to serve as a benchmark for evaluating text representation learning methods.  
% add some info on SNLI, also add MultiNLI
% Though, can we add MultiNLI? (because they don't provide the test set)

\paragraph{Semantic Textual Similarity}
While semantic relatedness requires training a model on top of the sentence embeddings, we also evaluate embeddings on the unsupervised SemEval tasks. These datasets include pairs of sentences taken from news articles, forum discussions, news conversations, headlines, image and video descriptions labeled with a similarity score between 0 and 5. The goal is to evaluate how the cosine distance between two sentences correlate with a human-labeled similarity score through Pearson and Spearman correlations. We include STS tasks from 2012 \cite{Agirre:2012semeval}, 2013\footnote{Due to License issues, we do not include the SMT subtask.} \cite{Agirre13sem2013}, 2014 \cite{agirrea2014semeval}, 2015 \cite{agirre2015semeval} and 2016 \cite{agirrea2016semeval}. Each of these tasks includes several subtasks. SentEval reports both the average and the weighted average (by number of samples in each subtask) of the Pearson and Spearman correlations. % Ranging from 2012-17, including the benchmark.. needs description and citations

\paragraph{Paraphrase detection}
The Microsoft Research Paraphrase Corpus (MRPC) \cite{Dolan:2004acl} is composed of pairs of sentences which have been extracted from news sources on the Web. Sentence pairs have been human-annotated according to whether they capture a paraphrase/semantic equivalence relationship. We use the same approach as with SICK-E, except that our classifier has only 2 classes, i.e., the aim is to predict whether the sentences are paraphrases or not. % The quantitative aspects of this are not in any of the tables

\paragraph{Caption-Image retrieval}
The caption-image retrieval task evaluates joint image and language feature models \cite{Lin:2014eccv}. The goal is either to rank a large collection of images by their relevance with respect to a given query caption (Image Retrieval), or ranking captions by their relevance for a given query image (Caption Retrieval). The COCO dataset provides a training set of 113k images with 5 captions each. The objective consists of learning a caption-image compatibility score $\mathcal{L}_{\text{cir}}(x,y)$ from a set of aligned image-caption pairs as training data. We use a pairwise ranking-loss $\mathcal{L}_{\text{cir}}(x,y)$: % The quantitative aspects of this are also not in any of the tables

\vspace{-0.5cm}
\begin{align*}
\sum_y \sum_k \max (0, \alpha - s(Vy, Ux) + s(Vy, Ux_k))\,+\\
\sum_x \sum_{k'} \max (0, \alpha - s(Ux, Vy) + s(Ux, Vy_{k'})),
\end{align*}
\vspace{-0.5cm}

where $(x,y)$ consists of an image $y$ with one of its associated captions $x$, $(y_{k})_{k}$ and $(y_{k'})_{k'}$ are negative examples of the ranking loss, $\alpha$ is the margin and $s$ corresponds to the cosine similarity. $U$ and $V$ are learned linear transformations that project the caption $x$ and the image $y$ to the same embedding space. We measure Recall@K, with K $\in \{1,5,10\}$, i.e., the percentage of images/captions for which the corresponding caption/image is one of the first K retrieved; and median rank. We use the same splits as \newcite{Karpathy:2015cvpr}, i.e., we use 113k images (each containing 5 captions) for training, 5k images for validation and 5k images for test. For evaluation, we split the 5k images in 5 random sets of 1k images on which we compute the mean $R@1$, $R@5$, $R@10$ and median (Med r) over the 5 splits. We include 2048-dimensional pretrained ResNet-101 \cite{He:2016cvpr} features for all images.
\section{Usage and Requirements}
\label{sec_usage}

\inserttransferclassif

Our evaluations comprise two different types: ones where we need to learn on top of the provided sentence representations (e.g. classification/regression) and ones where we simply take the cosine similarity between the two representations, as in the STS tasks. In the binary and multi-class classification tasks, we fit either a Logistic Regression classifier or an MLP with one hidden layer on top of the sentence representations. For the natural language inference tasks, where we are given two sentences $u$ and $v$, we provide the classifier with the input $\langle u, v, |u-v|, u*v \rangle$. To fit the Pytorch models, we use Adam \cite{Kingma:2014iclr}, with a batch size 64. We tune the L2 penalty of the classifier with grid-search on the validation set. When using Sent\-Eval, two functions should be implemented by the user: 

\begin{itemize}
\item \texttt{prepare(params, dataset)}: sees the whole dataset and applies any necessary preprocessing, such as constructing a lookup table of word embeddings (this function is optional); and
\item \texttt{batcher(params, batch)}: given a batch of input sentences, returns an array of the sentence embeddings for the respective inputs.
\end{itemize}

The main \texttt{batcher} function allows the user to encode text sentences using any Python framework. For example, the batcher function might be a wrapper around a model written in Pytorch, TensorFlow, Theano, DyNet, or any other framework\footnote{Or any other programming language, as long as the vectors can be passed to, or loaded from, code written in Python.}. To illustrate the use, here is an example of what an evaluation script looks like, having defined the prepare and batcher functions:

\begin{verbatim}
import senteval
se = senteval.engine.SE(
        params, batcher, prepare)
transfer_tasks = ['MR', 'CR']
results = se.eval(transfer_tasks)
\end{verbatim}

\paragraph{Parameters} Both functions make use of a \texttt{params} object, which contains the settings of the network and the evaluation. SentEval has several parameters that influence the evaluation procedure. These include the following:

\begin{itemize}
\item \texttt{task\_path} (str, required): path to the data.
\item \texttt{seed} (int): random seed for reproducibility.
\item \texttt{batch\_size} (int): size of minibatch of text sentences provided to batcher (sentences are sorted by length).
\item \texttt{kfold} (int): k in the kfold-validation (default: 10).
\end{itemize}

The default config is:
\begin{verbatim}
params = {'task_path': PATH_TO_DATA,
          'usepytorch': True,
          'kfold': 10}
\end{verbatim}

We also give the user the ability to customize the classifier used for the classification tasks.

\paragraph{Classifier}
To be comparable to the results published in the literature, users should use the following parameters for Logistic Regression:
\begin{verbatim}
params['classifier'] =
    {'nhid': 0, 'optim': 'adam',
     'batch_size': 64, 'tenacity': 5,
     'epoch_size': 4}
\end{verbatim}
The parameters of the classifier include:
\begin{itemize}
\item \texttt{nhid} (int): number of hidden units of the MLP; if nhid$>0$, a Multi-Layer Perceptron with one hidden layer and a Sigmoid nonlinearity is used.
\item \texttt{optim} (str): classifier optimizer (default: adam).
\item \texttt{batch\_size} (int): batch size for training the classifier (default: 64).
\item \texttt{tenacity} (int): stopping criterion; maximum number of times the validation error does not decrease.
\item \texttt{epoch\_size} (int): number of passes through the training set for one epoch.
\item \texttt{dropout} (float): dropout rate in the case of MLP.
\end{itemize}

For use cases where there are multiple calls to SentEval, e.g when evaluating the sentence encoder at every epoch of training, we propose the following prototyping set of parameters, which will lead to slightly worse results but will make the evaluation significantly faster:
\begin{verbatim}
params['classifier'] = 
    {'nhid': 0, 'optim': 'rmsprop',
     'batch_size': 128, 'tenacity': 3,
     'epoch_size': 2}
\end{verbatim}

You may also pass additional parameters to the \texttt{params} object in order which will further be accessible from the prepare and batcher functions (e.g a pretrained model).

\paragraph{Datasets} In order to obtain the data and preprocess it so that it can be fed into SentEval, we provide the \texttt{get\_transfer\_data.bash} script in the data directory. The script fetches the different datasets from their known locations, unpacks them and preprocesses them. We tokenize each of the datasets with the MOSES tokenizer \cite{Koehn:2007:MOS:1557769.1557821} and convert all files to UTF-8 encoding. Once this script has been executed, the task\_path parameter can be set to indicate the path of the data directory.

\paragraph{Requirements} SentEval is written in Python. In order to run the evaluations, the user will need to install numpy, scipy and recent versions of pytorch and scikit-learn. In order to facilitate research where no GPUs are available, we offer for the evaluations to be run on CPU (using scikit-learn) where possible. For the bigger datasets, where more complicated models are often required, for instance STS Benchmark, SNLI, SICK-R and the image-caption retrieval tasks, we recommend pytorch models on a single GPU.

\section{Baselines}
\label{sec_experiments}

Several baseline models are evaluated in Table \ref{table:classif_results}:

\begin{itemize}
\item Continuous bag-of-words embeddings (average of word vectors). We consider the most commonly used pretrained word vectors available, namely the fastText \cite{mikolov:lrec:2018} and the GloVe \cite{pennington2014glove} vectors trained on CommonCrawl.
\item SkipThought vectors \cite{ba2016layer}
\item InferSent vectors \cite{Conneau:2017emnlp}
\end{itemize}

\inserttransferrelated

In addition to these methods, we include the results of current state-of-the-art methods for which both the encoder and the classifier are trained on each task (no transfer). For GloVe and fastText bag-of-words representations, we report the results for Logistic Regression and Multi-Layer Perceptron (MLP). For the MLP classifier, we tune the dropout rate and the number of hidden units in addition to the L2 regularization. We do not observe any improvement over Logistic Regression for methods that already have a large embedding size (4096 for Infersent and 4800 for SkipThought). On most transfer tasks, supervised methods that are trained directly on each task still outperform transfer methods. Our hope is that SentEval will help the community build sentence representations with better generalization power that can outperform both the transfer and the supervised methods.

\section{Conclusion}
\label{sec_conclusion}

Universal sentence representations are a hot topic in NLP research. Making use of a generic sentence encoder allows models to generalize and transfer better, even when trained on relatively small datasets, which makes them highly desirable for downstream NLP tasks.

\-\hspace{0.1cm} We introduced SentEval as a fair, straightforward and centralized toolkit for evaluating sentence representations. We have aimed to make evaluation as easy as possible: sentence encoders can be evaluated by implementing a simple Python interface, and we provide a script to download the necessary evaluation datasets. In future work, we plan to enrich SentEval with additional tasks as the consensus on the best evaluation for sentence embeddings evolves. In particular, tasks that probe for specific linguistic properties of the sentence embeddings \cite{Shi:etal:2016,Adi:etal:2017} are interesting directions towards understanding how the encoder understands language. We hope that our toolkit will be used by the community in order to ensure that fully comparable results are published in research papers.
%Final version: emphasize stopping-wheel-reinvention and broad set of datasets more

% \section{Acknowledgements}

% \nocite{*}
\section{Bibliographical References}
\label{main:ref}

\bibliographystyle{lrec}
\bibliography{lrec}

\begin{thebibliography}{}

\bibitem[\protect\citename{Adi \bgroup et al.\egroup }2017]{Adi:etal:2017}
Adi, Y., Kermany, E., Belinkov, Y., Lavi, O., and Goldberg, Y.
\newblock (2017).
\newblock Fine-grained analysis of sentence embeddings using auxiliary
  prediction tasks.
\newblock In {\em Proceedings of ICLR Conference Track}, Toulon, France.
\newblock Published online:
  \url{https://openreview.net/group?id=ICLR.cc/2017/conference}.

\bibitem[\protect\citename{Agirre \bgroup et al.\egroup
  }2012]{Agirre:2012semeval}
Agirre, E., Diab, M., Cer, D., and Gonzalez-Agirre, A.
\newblock (2012).
\newblock Semeval-2012 task 6: A pilot on semantic textual similarity.
\newblock In {\em Proceedings of Semeval-2012}, pages 385--393.

\bibitem[\protect\citename{Agirre \bgroup et al.\egroup }2013]{Agirre13sem2013}
Agirre, E., Cer, D., Diab, M., Gonzalez-agirre, A., and Guo, W.
\newblock (2013).
\newblock sem 2013 shared task: Semantic textual similarity, including a pilot
  on typed-similarity.
\newblock In {\em In *SEM 2013: The Second Joint Conference on Lexical and
  Computational Semantics. Association for Computational Linguistics}.

\bibitem[\protect\citename{Agirre \bgroup et al.\egroup
  }2014]{agirrea2014semeval}
Agirre, E., Baneab, C., Cardiec, C., Cerd, D., Diabe, M., Gonzalez-Agirre, A.,
  Guof, W., Mihalceab, R., Rigaua, G., and Wiebeg, J.
\newblock (2014).
\newblock Semeval-2014 task 10: Multilingual semantic textual similarity.
\newblock {\em SemEval 2014}, page~81.

\bibitem[\protect\citename{Agirre \bgroup et al.\egroup
  }2015]{agirre2015semeval}
Agirre, E., Banea, C., Cardie, C., Cer, D.~M., Diab, M.~T., Gonzalez-Agirre,
  A., Guo, W., Lopez-Gazpio, I., Maritxalar, M., Mihalcea, R., et~al.
\newblock (2015).
\newblock Semeval-2015 task 2: Semantic textual similarity, english, spanish
  and pilot on interpretability.
\newblock In {\em SemEval@ NAACL-HLT}, pages 252--263.

\bibitem[\protect\citename{Agirre \bgroup et al.\egroup
  }2016]{agirrea2016semeval}
Agirre, E., Baneab, C., Cerd, D., Diabe, M., Gonzalez-Agirre, A., Mihalceab,
  R., Rigaua, G., Wiebef, J., and Donostia, B.~C.
\newblock (2016).
\newblock Semeval-2016 task 1: Semantic textual similarity, monolingual and
  cross-lingual evaluation.
\newblock {\em Proceedings of SemEval}, pages 497--511.

\bibitem[\protect\citename{Ba \bgroup et al.\egroup }2016]{ba2016layer}
Ba, J.~L., Kiros, J.~R., and Hinton, G.~E.
\newblock (2016).
\newblock Layer normalization.
\newblock {\em Advances in neural information processing systems (NIPS)}.

\bibitem[\protect\citename{Bowman \bgroup et al.\egroup
  }2015]{Bowman:2015emnlp}
Bowman, S.~R., Angeli, G., Potts, C., and Manning, C.~D.
\newblock (2015).
\newblock A large annotated corpus for learning natural language inference.
\newblock In {\em Proceedings of EMNLP}.

\bibitem[\protect\citename{Cer \bgroup et al.\egroup }2017]{cer2017semeval}
Cer, D., Diab, M., Agirre, E., Lopez-Gazpio, I., and Specia, L.
\newblock (2017).
\newblock Semeval-2017 task 1: Semantic textual similarity-multilingual and
  cross-lingual focused evaluation.
\newblock {\em arXiv preprint arXiv:1708.00055}.

\bibitem[\protect\citename{Chiu \bgroup et al.\egroup }2016]{Chiu:2016repeval}
Chiu, B., Korhonen, A., and Pyysalo, S.
\newblock (2016).
\newblock Intrinsic evaluation of word vectors fails to predict extrinsic
  performance.
\newblock In {\em First Workshop on Evaluating Vector Space Representations for
  NLP (RepEval)}.

\bibitem[\protect\citename{Conneau \bgroup et al.\egroup
  }2017]{Conneau:2017emnlp}
Conneau, A., Kiela, D., Schwenk, H., Barrault, L., and Bordes, A.
\newblock (2017).
\newblock Supervised learning of universal sentence representations from
  natural language inference data.
\newblock In {\em Proceedings of EMNLP}, Copenhagen, Denmark.

\bibitem[\protect\citename{Dolan \bgroup et al.\egroup }2004]{Dolan:2004acl}
Dolan, B., Quirk, C., and Brockett, C.
\newblock (2004).
\newblock Unsupervised construction of large paraphrase corpora: Exploiting
  massively parallel news sources.
\newblock In {\em Proceedings of ACL}, page 350.

\bibitem[\protect\citename{Ettinger \bgroup et al.\egroup
  }2016]{Ettinger:2016acl}
Ettinger, A., Elgohary, A., and Resnik, P.
\newblock (2016).
\newblock Probing for semantic evidence of composition by means of simple
  classification tasks.
\newblock In {\em First Workshop on Evaluating Vector Space Representations for
  NLP (RepEval)}, page 134.

\bibitem[\protect\citename{Faruqui \bgroup et al.\egroup
  }2016]{Faruqui:2016repeval}
Faruqui, M., Tsvetkov, Y., Rastogi, P., and Dyer, C.
\newblock (2016).
\newblock Problems with evaluation of word embeddings using word similarity
  tasks.
\newblock {\em arXiv preprint arXiv:1605.02276}.

\bibitem[\protect\citename{He \bgroup et al.\egroup }2016]{He:2016cvpr}
He, K., Zhang, X., Ren, S., and Sun, J.
\newblock (2016).
\newblock Deep residual learning for image recognition.
\newblock In {\em Proceedings of CVPR}.

\bibitem[\protect\citename{Hill \bgroup et al.\egroup }2016a]{Hill:2016naacl}
Hill, F., Cho, K., and Korhonen, A.
\newblock (2016a).
\newblock Learning distributed representations of sentences from unlabelled
  data.
\newblock In {\em Proceedings of NAACL}.

\bibitem[\protect\citename{Hill \bgroup et al.\egroup }2016b]{Hill:2016cl}
Hill, F., Reichart, R., and Korhonen, A.
\newblock (2016b).
\newblock Simlex-999: Evaluating semantic models with (genuine) similarity
  estimation.
\newblock {\em Computational Linguistics}.

\bibitem[\protect\citename{Hu and Liu}2004]{Hu:2004kdd}
Hu, M. and Liu, B.
\newblock (2004).
\newblock Mining and summarizing customer reviews.
\newblock In {\em Proceedings of SIGKDD}, pages 168--177.

\bibitem[\protect\citename{Ji and Eisenstein}2013]{ji2013discriminative}
Ji, Y. and Eisenstein, J.
\newblock (2013).
\newblock Discriminative improvements to distributional sentence similarity.
\newblock In {\em Proceedings of the 2013 Conference on Empirical Methods in
  Natural Language Processing (EMNLP)}.

\bibitem[\protect\citename{Karpathy and Fei-Fei}2015]{Karpathy:2015cvpr}
Karpathy, A. and Fei-Fei, L.
\newblock (2015).
\newblock Deep visual-semantic alignments for generating image descriptions.
\newblock In {\em Proceedings of CVPR}, pages 3128--3137.

\bibitem[\protect\citename{Kingma and Ba}2014]{Kingma:2014iclr}
Kingma, D.~P. and Ba, J.
\newblock (2014).
\newblock Adam: A method for stochastic optimization.
\newblock In {\em Proceedings of the 3rd International Conference on Learning
  Representations (ICLR)}.

\bibitem[\protect\citename{Kiros \bgroup et al.\egroup }2015]{Kiros:2015nips}
Kiros, R., Zhu, Y., Salakhutdinov, R.~R., Zemel, R., Urtasun, R., Torralba, A.,
  and Fidler, S.
\newblock (2015).
\newblock Skip-thought vectors.
\newblock In {\em Advances in neural information processing systems}, pages
  3294--3302.

\bibitem[\protect\citename{Koehn \bgroup et al.\egroup
  }2007]{Koehn:2007:MOS:1557769.1557821}
Koehn, P., Hoang, H., Birch, A., Callison-Burch, C., Federico, M., Bertoldi,
  N., Cowan, B., Shen, W., Moran, C., Zens, R., Dyer, C., Bojar, O.,
  Constantin, A., and Herbst, E.
\newblock (2007).
\newblock Moses: Open source toolkit for statistical machine translation.
\newblock In {\em Proceedings of the 45th Annual Meeting of the ACL on
  Interactive Poster and Demonstration Sessions}, ACL '07, pages 177--180,
  Stroudsburg, PA, USA. Association for Computational Linguistics.

\bibitem[\protect\citename{Lai and Hockenmaier}2014]{lai2014illinois}
Lai, A. and Hockenmaier, J.
\newblock (2014).
\newblock Illinois-lh: A denotational and distributional approach to semantics.
\newblock {\em Proc. SemEval}, 2:5.

\bibitem[\protect\citename{Lin \bgroup et al.\egroup }2014]{Lin:2014eccv}
Lin, T.-Y., Maire, M., Belongie, S., Hays, J., Perona, P., Ramanan, D.,
  Doll{\'a}r, P., and Zitnick, C.~L.
\newblock (2014).
\newblock Microsoft coco: Common objects in context.
\newblock In {\em European Conference on Computer Vision}, pages 740--755.
  Springer International Publishing.

\bibitem[\protect\citename{Marelli \bgroup et al.\egroup
  }2014]{Marelli:2014lrec}
Marelli, M., Menini, S., Baroni, M., Bentivogli, L., Bernardi, R., and
  Zamparelli, R.
\newblock (2014).
\newblock A {SICK} cure for the evaluation of compositional distributional
  semantic models.
\newblock In {\em Proceedings of LREC}.

\bibitem[\protect\citename{Mikolov \bgroup et al.\egroup
  }2017]{mikolov:lrec:2018}
Mikolov, T., Grave, E., Bojanowski, P., Puhrsch, C., and Joulin, A.
\newblock (2017).
\newblock Advances in pre-training distributed word representations.

\bibitem[\protect\citename{Nayak \bgroup et al.\egroup }2016]{Nayak:2016acl}
Nayak, N., Angeli, G., and Manning, C.~D.
\newblock (2016).
\newblock Evaluating word embeddings using a representative suite of practical
  tasks.
\newblock In {\em First Workshop on Evaluating Vector Space Representations for
  NLP (RepEval)}, page~19.

\bibitem[\protect\citename{Pang and Lee}2004]{Pang:2004acl}
Pang, B. and Lee, L.
\newblock (2004).
\newblock A sentimental education: Sentiment analysis using subjectivity
  summarization based on minimum cuts.
\newblock In {\em Proceedings of ACL}, page 271.

\bibitem[\protect\citename{Pang and Lee}2005]{Pang:2005acl}
Pang, B. and Lee, L.
\newblock (2005).
\newblock Seeing stars: Exploiting class relationships for sentiment
  categorization with respect to rating scales.
\newblock In {\em Proceedings of ACL}, pages 115--124.

\bibitem[\protect\citename{Pennington \bgroup et al.\egroup
  }2014]{pennington2014glove}
Pennington, J., Socher, R., and Manning, C.~D.
\newblock (2014).
\newblock Glove: Global vectors for word representation.
\newblock In {\em Proceedings of the 2014 Conference on Empirical Methods in
  Natural Language Processing (EMNLP)}, volume~14, pages 1532--1543.

\bibitem[\protect\citename{Shi \bgroup et al.\egroup }2016]{Shi:etal:2016}
Shi, X., Padhi, I., and Knight, K.
\newblock (2016).
\newblock Does string-based neural {MT} learn source syntax?
\newblock In {\em Proceedings of EMNLP}, pages 1526--1534, Austin, Texas.

\bibitem[\protect\citename{Socher \bgroup et al.\egroup
  }2013]{Socher:2013emnlp}
Socher, R., Perelygin, A., Wu, J.~Y., Chuang, J., Manning, C.~D., Ng, A.~Y.,
  Potts, C., et~al.
\newblock (2013).
\newblock Recursive deep models for semantic compositionality over a sentiment
  treebank.
\newblock In {\em Proceedings of EMNLP}, pages 1631---1642.

\bibitem[\protect\citename{Tai \bgroup et al.\egroup }2015a]{Tai:2015acl}
Tai, K.~S., Socher, R., and Manning, C.~D.
\newblock (2015a).
\newblock Improved semantic representations from tree-structured long
  short-term memory networks.
\newblock {\em Proceedings of ACL}.

\bibitem[\protect\citename{Tai \bgroup et al.\egroup }2015b]{tai2015improved}
Tai, K.~S., Socher, R., and Manning, C.~D.
\newblock (2015b).
\newblock Improved semantic representations from tree-structured long
  short-term memory networks.
\newblock {\em Proceedings of the 53rd Annual Meeting of the Association for
  Computational Linguistics (ACL)}.

\bibitem[\protect\citename{Voorhees and Tice}2000]{Voorhees:2000sigir}
Voorhees, E.~M. and Tice, D.~M.
\newblock (2000).
\newblock Building a question answering test collection.
\newblock In {\em Proceedings of the 23rd annual international ACM SIGIR
  conference on Research and development in information retrieval}, pages
  200--207. ACM.

\bibitem[\protect\citename{Vuli{\'c} \bgroup et al.\egroup
  }2017]{Vulic:2017eacl}
Vuli{\'c}, I., Kiela, D., and Korhonen, A.
\newblock (2017).
\newblock Evaluation by association: A systematic study of quantitative word
  association evaluation.
\newblock In {\em Proceedings of EACL}, volume~1, pages 163--175.

\bibitem[\protect\citename{Wiebe \bgroup et al.\egroup }2005]{Wiebe:2005lrec}
Wiebe, J., Wilson, T., and Cardie, C.
\newblock (2005).
\newblock Annotating expressions of opinions and emotions in language.
\newblock {\em Language resources and evaluation}, 39(2):165--210.

\bibitem[\protect\citename{Wieting \bgroup et al.\egroup
  }2015]{wieting2015towards}
Wieting, J., Bansal, M., Gimpel, K., and Livescu, K.
\newblock (2015).
\newblock Towards universal paraphrastic sentence embeddings.
\newblock {\em Proceedings of the 4th International Conference on Learning
  Representations (ICLR)}.

\bibitem[\protect\citename{Wieting \bgroup et al.\egroup
  }2016]{wieting2016charagram}
Wieting, J., Bansal, M., Gimpel, K., and Livescu, K.
\newblock (2016).
\newblock Charagram: Embedding words and sentences via character n-grams.
\newblock {\em Proceedings of the 2016 Conference on Empirical Methods in
  Natural Language Processing (EMNLP)}.

\bibitem[\protect\citename{Zhao \bgroup et al.\egroup }2015]{zhao2015self}
Zhao, H., Lu, Z., and Poupart, P.
\newblock (2015).
\newblock Self-adaptive hierarchical sentence model.
\newblock In {\em Proceedings of the 24th International Conference on
  Artificial Intelligence}, IJCAI'15, pages 4069--4076. AAAI Press.

\bibitem[\protect\citename{Zhou \bgroup et al.\egroup }2016]{zhou2016text}
Zhou, P., Qi, Z., Zheng, S., Xu, J., Bao, H., and Xu, B.
\newblock (2016).
\newblock Text classification improved by integrating bidirectional lstm with
  two-dimensional max pooling.
\newblock {\em Proceedings of COLING 2016, the 26th International Conference on
  Computational Linguistics}.

\end{thebibliography}

%\section{Language Resource References}
%\label{lr:ref}
%\bibliographystylelanguageresource{lrec}
%\bibliographylanguageresource{lrec}

\end{document}